\documentclass[runningheads]{llncs}

\usepackage{graphicx}
\usepackage[table,xcdraw]{xcolor}
\usepackage{cite}
\usepackage[toc,page]{appendix}
\usepackage{caption}
\usepackage{floatrow} 
\newfloatcommand{capbtabbox}{table}[][\FBwidth] 
\usepackage[table,xcdraw]{xcolor}
\usepackage{blindtext} 
\usepackage{subcaption}
\begin{document}
\title{Improving Prediction of Low-Prior Clinical Events with Simultaneous General Patient-State Representation Learning}
\titlerunning{Simultaneous Low-Prior Event and GPSR Learning}
%
\author{Matthew Barren\orcidID{0000-0003-0855-2144} \and 
Milos Hauskrecht\orcidID{0000-0002-7818-0633}}
\authorrunning{M. Barren and M. Hauskrecht}

\institute{University of Pittsburgh, Pittsburgh PA 15260, USA 
\email{\{mpb43, milos\}@pitt.edu}}
\vspace{-4mm}
\maketitle              
\vspace{-4mm}
\begin{abstract}
\vspace{-4mm}
Low-prior targets are common among many important clinical events, which introduces the challenge of having enough data to support learning of their predictive models. Many prior works have addressed this problem by first building a general patient-state representation model, and then adapting it to a new low-prior prediction target. In this schema, there is potential for the predictive performance to be hindered by the misalignment between the general patient-state model and the target task. To overcome this challenge, we propose a new method that simultaneously optimizes a shared model through multi-task learning of both the low-prior supervised target and general purpose patient-state representation (GPSR). More specifically, our method improves prediction performance of a low-prior task by jointly optimizing a shared model that combines the loss of the target event and a broad range of generic clinical events. We study the approach in the context of Recurrent Neural Networks (RNNs). Through extensive experiments on multiple clinical event targets using MIMIC-III \cite{johnson2016mimic} data, we show that the inclusion of general patient-state representation tasks during model training improves the prediction of individual low-prior targets.
\vspace{-2mm}
\end{abstract}
\vspace{-4mm}
\vspace{-4mm}
\section{Introduction}
\vspace{-4mm}
Across machine learning domains, many important events are difficult to predict because of their low-prior probability. This situation is frequent in clinical event prediction, where severe events and interventions are both uncommon and imperative to foresee. To some degree, low-priors are a constraint of the task definition. For example, the prediction of first sepsis onset will have at most one positive instance per a patient hospitalization, and therefore it is constrained by design. Additionally, in temporal modeling this prior is further reduced by the frequency (e.g. predict every two hours) and time horizon of prediction.

Previous machine learning works have utilized general patient-state representations (GPSRs) \cite{miotto2016deep, gupta2018using, lei2018effective, lyu2018improving, rajkomar2018scalable} or transfer learning \cite{gupta2020transfer} as methods to deal with low-prior events. However, in both cases, there is a potential for predictive performance to be hindered by the misalignment of the general purpose model and the low-prior target. For example, in GPSR learning, it is possible that the extracted representation features obfuscate the signals from the raw inputs that are highly important for accurately predicting a septic patient. 

In order to improve the prediction performance of low-prior clinical events, we propose a new method that simultaneously trains a shared model that can support both low-prior target prediction and general patient-state representation tasks. Accordingly, the parameters of the model are optimized through a two-component loss function. To better tune the model to the desired low-prior clinical task, a weight parameter is used to adjust the influence between the low-prior target and GPSR. Thus, a GPSR is learned jointly to aid a specific prediction target instead of being used as an upstream step to accommodate it. We explore our method in the context of recurrent neural networks (RNNs) with long short-term memory cells (LSTM). LSTMs have been used to define both a GPSR models for clinical sequences \cite{lyu2018improving, lee2021modeling}, as well as, a model for predicting single events from past clinical sequences \cite{tomavsev2019clinically}. 

We explore the benefits of our simultaneous learning method experimentally using clinical data derived from the MIMIC-III database \cite{johnson2016mimic} predicting three low-prior events: 72 hour mortality, 6 hour sepsis onset, and 2 hour norepinephrine administration. These targets have priors that range from 0.0013 to 0.0109. The GPSR component of the LSTM model is defined as a broad range of clinical lab and vital sign events that are one-hot encoded to normal and abnormal values. Through extensive experiments we show that simultaneously optimizing the GPSR and the low-prior prediction task leads to models with improved prediction performance as measured by the area under the precision-recall curve (AUPRC). In addition, two ablation studies reducing the event priors and samples in the training data demonstrate the robustness of our approach.
\vspace{-4mm}
\section{Related Work}
\vspace{-4mm}
{\bf General patient-state representation learning.}  General patient-state representations are often desirable for their ability to compress complex data into a lower dimensional representation with the goal of accurately representing the signals that are inherent to the patient-state. General patient-state representation models include a wide range of standard matrix factorization approaches and modern neural architectures models. Examples of GPSR models include Singular Value Decomposition (SVD) \cite{malakouti2019predicting}, autoencoder architectures \cite{miotto2016deep, rajkomar2018scalable}, recurrent neural network models \cite{gupta2018using, lei2018effective,lee2021modeling}, attention mechanisms \cite{choi2016retain}, and composites of the previously mentioned paradigms \cite{lyu2018improving}.  For example, the authors of DeepPatient used a denoising autoencoder (AE) to learn a patient representation over time windows of clinical observations and applied it predict patient diseases. The concept of autoencoding has further been applied in the space of sequence models, such as LSTMs \cite{gupta2018using, lei2018effective, lyu2018improving}.

{\bf Task specific model learning.}  Task specific model learning optimizes parameters based only on a supervised target(s). Models are trained using a supervised loss. Especially suitable for this purpose, are autoregressive models that provide an end-to-end framework for defining the model, inputs, outputs, and target task loss. LSTMs applied to clinical tasks have been found to provide strong predictive performance, such as predicting chronic kidney injury \cite{tomavsev2019clinically}.

{\bf Simultaneous GPSR and tasks-specific learning.} Simultaneous learning of two different task paradigms has been explored in the space of topic models. Supervised Latent Dirichlet Allocation (sLDA) combines the objective function of the expectation of a token belonging to a topic with a supervised task to guide topic learning \cite{blei2010supervised}. In prediction focused sLDA (pf-sLDA) \cite{ren2019prediction}, the authors used weighting to examine the balance between topic and supervised tasks. In our work, we take a similar approach to pf-sLDA by jointly modeling a supervised task with a general patient representation. In contrast to the work of Ren et. al., we solely focus on the performance of the supervised task, and use the general patient representation as support to improve prediction performance.
\vspace{-4mm}
\section{Methodology}
\vspace{-2mm}
\subsection{Model Definition}
\vspace{-2mm}
Our objective is to learn a model $f: X \rightarrow Y$ that can predict a future target event $Y$ from past observations $X$. Since past observations grow in time, $X$ is often replaced with a fixed length summary vector $S$. A summary vector can be developed from a number of different strategies, such as using feature templates that featurize time-series of all clinical variables defining $X$ \cite{hauskrecht2013outlier, hauskrecht2016outlier} or by compressing observations to a low-dimensional space using SVD \cite{malakouti2019predicting}. RNNs have had success with learning clinical targets by segmenting past observations $X$ to the current time, $t$, into sequences of observations $X_1, X_2, \dots, X_t$. A summary state is then defined as $S_t$ representing the hidden state of the RNN. In this work, we consider an RNN model with Long Short-Term Memory units \cite{gers2000learning} to represent the predictive model $f$.

\vspace{-4mm}
\begin{figure}
\centering
\includegraphics[scale=0.25]{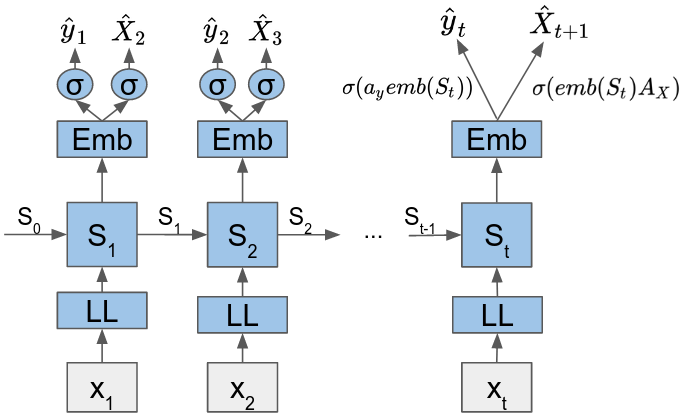}
\caption{The shared low-prior clinical prediction task and GPSR task architecture. LL in the model architecture stands for linear layer.}
\label{fig:proposed_model}
\end{figure}
\vspace{-6mm}
Since learning a model $f$ for a low-prior target is sensitive to the training data size, we propose to aid its learning with a GPSR model. A GPSR model is not tailored to cover one specific prediction task, instead it tries to represent the overall data sequences and their characteristics. In this work, we consider an LSTM-based model of GPSRs. This GPSR is defined to predict future clinical observations based on the past events, $g: X_1, X_2, \dots, X_t \rightarrow X_{t+1}$. Each one-hot encoded generic clinical task in $X_{t+1}$ can be represented as a set of multi-class targets $R = (r_1, r_2, ..., r_d)$, where each clinical event, $r_i$, is a set of discretized class values for that clinical event, $r_i=(c_1, c_2, ..., c_m)$. The reason for choosing a GPSR LSTM-based model is that it can be aligned with the LSTM event prediction model $f$. Briefly, the GPSR LSTM model can be summarized in terms of a state, $S_t$, similarly to the LSTM model of the model $f$, but instead of just one target event, it predicts a broad range of clinical events.

The key idea of our approach is to have $f$ and $g$ share parameters to learn both tasks simultaneously. More specifically, $f$ and $g$ can be defined on the same summary state, $S_t$, and thus can be redefined to $f': S_{t} \rightarrow Y_{t}$ and $g': S_{t} \rightarrow X_{t+1}$. The summary state can further be defined as a function of the current input and past state, $m: X_t, S_{t-1} \rightarrow  S_{t}$. In our model a shared embedding layer, $emb(S_t)$, is applied to the summary state $S_t$ before prediction outputs are computed. The $emb(S_t)$ layer is a linear layer with a rectified linear unit (ReLU) activation. Thus the current state can be computed from eq. \ref{eq: St}, which is a function of the shared architecture between the low-prior target and the GPSR. The sigmoid of the dot product between the target weight vector,  $a_{y}$, and $S_t$ are computed to get the low-prior prediction, eq. \ref{eq: target_pred}. Similarly, the next-step patient representation can be computed from the sigmoid of the state vector $S_t$ being multiplied to the GPSR weight matrix, $A_X$,  eq. \ref{eq: gpsr_pred}. Figure \ref{fig:proposed_model} graphically depicts the described model definition.

\vspace{-4mm}
\begin{equation}
S_{t} = m(X_t, S_{t-1})
\label{eq: St}
\end{equation}
\vspace{-6mm}
\begin{equation}
\hat{y}_{t} = f'(S_t) = \sigma(a_{y}emb(S_t))
\label{eq: target_pred}
\end{equation}
\vspace{-6mm}
\begin{equation}
\hat{X}_{t+1} = g'(S_t) = \sigma(emb(S_t)A_{X})
\label{eq: gpsr_pred}
\end{equation}

Further, the errors of the prediction target and GPSR tasks can be computed using cross entropy in eqs. \ref{eq: target error} and \ref{eq: task error}. Eq. \ref{eq: task error} computes the error for each GPSR task, $r$, in $X$.
\vspace{-4mm}
\begin{equation}
err_{y}(\hat{y_{t}}, y_{t}) = -[y_t\log(\hat{y}_t) + (1-y_t)\log(1-\hat{y}_t)]
\label{eq: target error}
\end{equation}
\vspace{-6mm}
\begin{equation}
err^{r}_{X}(\hat{X}^{r}_{t+1}, X^{r}_{t+1}) = -\Sigma_{c}^{r} X_{t+1}^{r,c}\log(\hat{X}_{t+1}^{r,c})
\label{eq: task error}
\end{equation}
\vspace{-6mm}
\subsection{Optimizing Weighted Simultaneous Learning Loss}
\vspace{-1mm}
Similar to Ren et. al.\cite{ren2019prediction}, weighting is applied to control the parameter learning, weights and LSTM gates, influence between the two objectives, $f'$ and $g'$. The loss function, eq. \ref{loss fn}, uses a hyperparameter, $p$, to weight the errors between the low-prior target and the GPSR tasks. Setting $p=1$ results in a low-prior event only driven model, and conversely, $p=0$ will yield parameter updates based only on the GPSR tasks. For optimizing model weights, Adaptive moment with decoupled weight decay was used (AdamW)\cite{loshchilov2017decoupled}.
\vspace{-2mm}
\begin{equation}
l(X_{t+1}) = p(err_{y}(\hat{y}_{t}, y_{t})) + (1-p)\frac{\Sigma_{r}^{R} err^{r}_{X}(\hat{X}^{r}_{t+1}, X^{r}_{t+1})}{|R|}
\label{loss fn}
\end{equation}
\vspace{-6mm}
\section{Experiments}
\vspace{-2mm}
\subsection{Simultaneous Model Architectures}
\vspace{-1mm}
In this paper, two simultaneous model architectures are proposed. A third was trained/evaluated, but excluded from the results due to it's similar performance to \textbf{Evt+GPSR} model. They each use the same base architecture shown in Figure \ref{fig:proposed_model}, but the latter model extends the network with additional layers that are task specific. The proposed models are the following:
\vspace{-2mm}
\begin{itemize}
\item Low-Prior Event Target and GPSR (\textbf{Evt+GPSR}) [Figure \ref{fig:proposed_model}]
\item Low-Prior Event with Linear Layer and GPSR with Multi-task Linear Layer (\textbf{EvtLL+GPSR-MTLL}) [Figure  \ref{fig:evtLL_GPSRLL}]
\end{itemize}
\vspace{-2mm}
\vspace{-4mm}
\subsection{Baseline Model Architectures}
\begin{figure}[hbt!]
	\centering
	\begin{subfigure}[b]{0.13\textwidth}
	\centering
	\includegraphics[width=\textwidth]{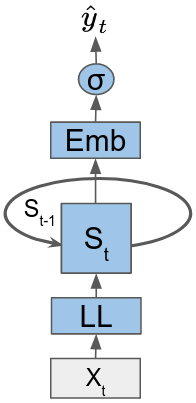}
	\caption{•}
	\label{fig:supervised}
	\end{subfigure}
	\hfill
	\begin{subfigure}[b]{0.27\textwidth}
	\centering
	\includegraphics[width=\textwidth]{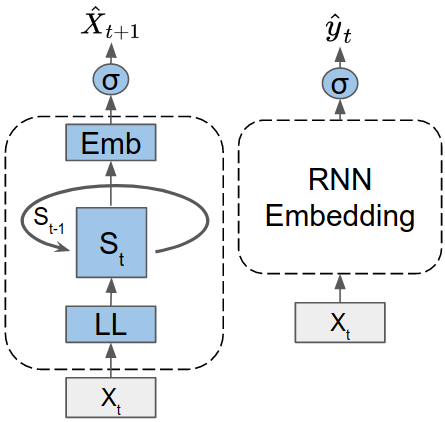}
	\caption{•}
	\label{fig:embedding}
	\end{subfigure}
	\hfill
	\begin{subfigure}[b]{0.25\textwidth}
	\centering
	\includegraphics[width=\textwidth]{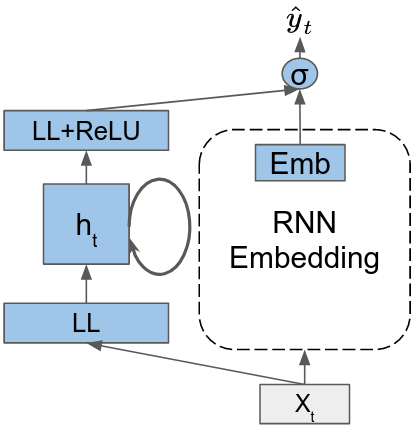}
	\caption{•}
	\label{fig:residual}
	\end{subfigure}
	\hspace{5mm}
	\begin{subfigure}[b]{0.27\textwidth}
	\centering
	\includegraphics[width=\textwidth]{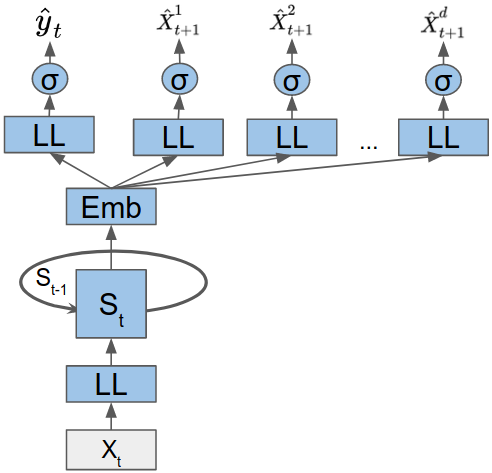}
	\caption{•}
	\label{fig:evtLL_GPSRLL}
	\end{subfigure}
	\vspace{-3mm}
	\caption{\textbf{Figures \ref{fig:supervised}, \ref{fig:embedding}, and \ref{fig:residual}} are the baseline prediction models, \textbf{RNN Spv.}, \textbf{RNN Embedding}, and \textbf{RNN Residual}. Likewise, \ref{fig:evtLL_GPSRLL} is the proposed model with an extended structure, \textbf{RNN EvtLL+GPSR-MTLL}}
\end{figure}

\vspace{-2mm}
Three baseline models are used to compare with the proposed simultaneous models. The same general model structure given in Figure \ref{fig:proposed_model} is used for each with some modifications to their respective prediction objective. 
\vspace{-2mm}
\begin{itemize}
\item \textbf{Supervised model (RNN Spv.)} low-prior task-specific model (Figure \ref{fig:supervised})

\item \textbf{RNN Embedding} is a GPSR model that is trained to forecast the generic clinical events. The supervised target is then learned using a single linear layer based on the features learned from the embedding model. For the experiments, this model is trained to each prediction time horizon to align with the low-prior prediction targets. [Figure \ref{fig:embedding}]

\item \textbf{RNN Residual} uses the learned RNN Embedding model with additional residual layers. The learned embedding layer continues to be optimized during supervised training to allow for additional tuning to the target. The residual layers attempt to learn the low-prior target signal from raw inputs that were not captured in the GPSR embedding model. [Figure \ref{fig:residual}]
\end{itemize}
Thus, \textbf{RNN Spv.} is equivalent to \textbf{Evt+GPSR} if the loss weight value is set equal to 1.0, $p=1$, where only the supervised low-prior task influences parameter learning. Likewise, the other baselines utilize GPSR learning in a sequential fashion where the target is not simultaneously considered with the GPSR tasks.

\vspace{-6mm}
\vspace{-10mm}
\subsection{Low-Prior Targets}
\vspace{-2mm}
Experiments were conducted using MIMIC-III's\cite{johnson2016mimic} electronic healthcare record data set. The data set included ICU patients of 18 years of age and older with an inpatient time that exceeds both 24 hours and the prediction horizon. Target statistics can be found in Table \ref{tbl:dataset info}. The prediction time horizons for mortality, sepsis, and intravenous (IV) norepinephrine were 72, 6, and 2 hours, respectively. The separation time between each instance in a sequence is the same as the prediction horizon except for the mortality task, which uses 24 hour sequence intervals. Sepsis prediction targets were generated based on Physionet's competition \cite{reyna2019early}. The IV Norepinephrine task is a prediction of a new medication administration. A patient may have multiple administrations in a single hospital stay, and to determine a new delivery, the half-life of the medication was compared to statistics of subsequent drug time intervals. Given the short half-life of 2.5 minutes \cite{smith2019norepinephrine} and the distribution of subsequent administration intervals, a holdout period of 2 hours after drug delivery is applied before predictions may resume (i.e during drug administration, prediction is suspended).
\vspace{-4mm}
\begin{figure}[hbt!]
\begin{floatrow}
\scriptsize
\centering
\capbtabbox[12cm]{%
\begin{tabular}{lcccccccccccc}
 &
  \multicolumn{12}{c}{\textbf{Data Sets}} \\ \cline{2-13} 
\multicolumn{1}{l|}{} &
  \multicolumn{4}{c|}{\cellcolor[HTML]{C0C0C0}\textbf{Mortality}} &
  \multicolumn{4}{c}{\cellcolor[HTML]{C0C0C0}\textbf{Norepinephrine}} &
  \multicolumn{4}{c|}{\cellcolor[HTML]{C0C0C0}\textbf{Sepsis}} \\ \cline{2-13} 
\multicolumn{1}{l|}{} &
  \textbf{Adms} &
  \textbf{\# Pos} &
  \textbf{\# Neg} &
  \multicolumn{1}{c|}{\textbf{Prior}} &
  \textbf{Adms} &
  \textbf{\# Pos} &
  \textbf{\# Neg} &
  \multicolumn{1}{c|}{\textbf{Prior}} &
  \textbf{Adms} &
  \textbf{\# Pos} &
  \textbf{\# Neg} &
  \multicolumn{1}{c|}{\textbf{Prior}} \\ \cline{2-13} 
\multicolumn{1}{c|}{Train} &
  8803 &
  700 &
  82,797 &
  \multicolumn{1}{c|}{0.0084} &
  11,694 &
  948 &
  1,156,503 &
  \multicolumn{1}{c|}{0.0008} &
  11,694 &
  506 &
  384,097 &
  \multicolumn{1}{c|}{0.0013} \\ \hline
\multicolumn{1}{c|}{Valid} &
  2363 &
  214 &
  18,544 &
  \multicolumn{1}{c|}{0.0114} &
  3,141 &
  370 &
  248,929 &
  \multicolumn{1}{c|}{0.0013} &
  3141 &
  140 &
  90,687 &
  \multicolumn{1}{c|}{0.0015} \\ \hline
\multicolumn{1}{c|}{Test} &
  4850 &
  425 &
  38,626 &
  \multicolumn{1}{c|}{0.0109} &
  6,535 &
  750 &
  589,720 &
  \multicolumn{1}{c|}{0.0013} &
  6,535 &
  287 &
  189,092 &
  \multicolumn{1}{c|}{0.0015} \\ \cline{2-13} 
\end{tabular}
}%
{
\vspace{-4mm}
\caption{Data set statistics for each low-prior event.}
\label{tbl:dataset info}
}
\end{floatrow}
\end{figure}
\vspace{-12mm}
\subsection{Inputs and GPSR Tasks}
\vspace{-2mm}
The inputs for each task are 191 lab and vital sign observations that are discretized to a one-hot encoding of normal/abnormal or normal/abnormal low/abnormal high. This discretization is based on a knowledge base of normal ranges that were compiled from \cite{kratz2004laboratory,laposata2019laposata, mcdonald2003loinc}. A last value carry forward (LVCF) method is applied for each observation relative to the prediction point, and in the event no prior observation exists, the encoding positions for that observation remain zero. 

The generic clinical tasks for the GPSR were 189 of the 191 laboratory and vital sign observations. Two observations were excluded as a task because their presence by definition is abnormal. A LVCF was also used for the target classes, and in the event that no value exists, a normal class is imputed. For this paper, the time horizons of the generic clinical tasks aligned with the respective prediction target time. For example the 6 hour sepsis target had GPSR tasks of 6 hour time horizons too. This was done for all models that utilized generic clinical tasks for their GPSR.
\vspace{-2mm}
\subsection{Model Training and Selection}
\vspace{-1mm}
Since AUPRC is the primary metric for evaluating low-prior event performance, AUROC was used to determine early stopping to avoid biasing model selection on a single evaluation metric. All models were trained over a number of epochs with early stopping based on validation AUROC. Dropout was applied for the linear layers. For each model architecture the same set of layer sizes were explored along with regularization parameters on the supervised output parameters over multiple iterations. For example, both \textbf{RNN Spv.} and \textbf{Evt+GPSR} explored the same set of layer configurations. The best performing average validation AUROC determined the model hyperparameters. The GPSR embedding model was trained with early stopping based on the tolerance of the validation loss.
\vspace{-4mm}
\subsection{Weighted Loss Selection}
\begin{figure}
\begin{floatrow}
\tiny
\ffigbox[5cm]{%
	\includegraphics[scale=0.3]{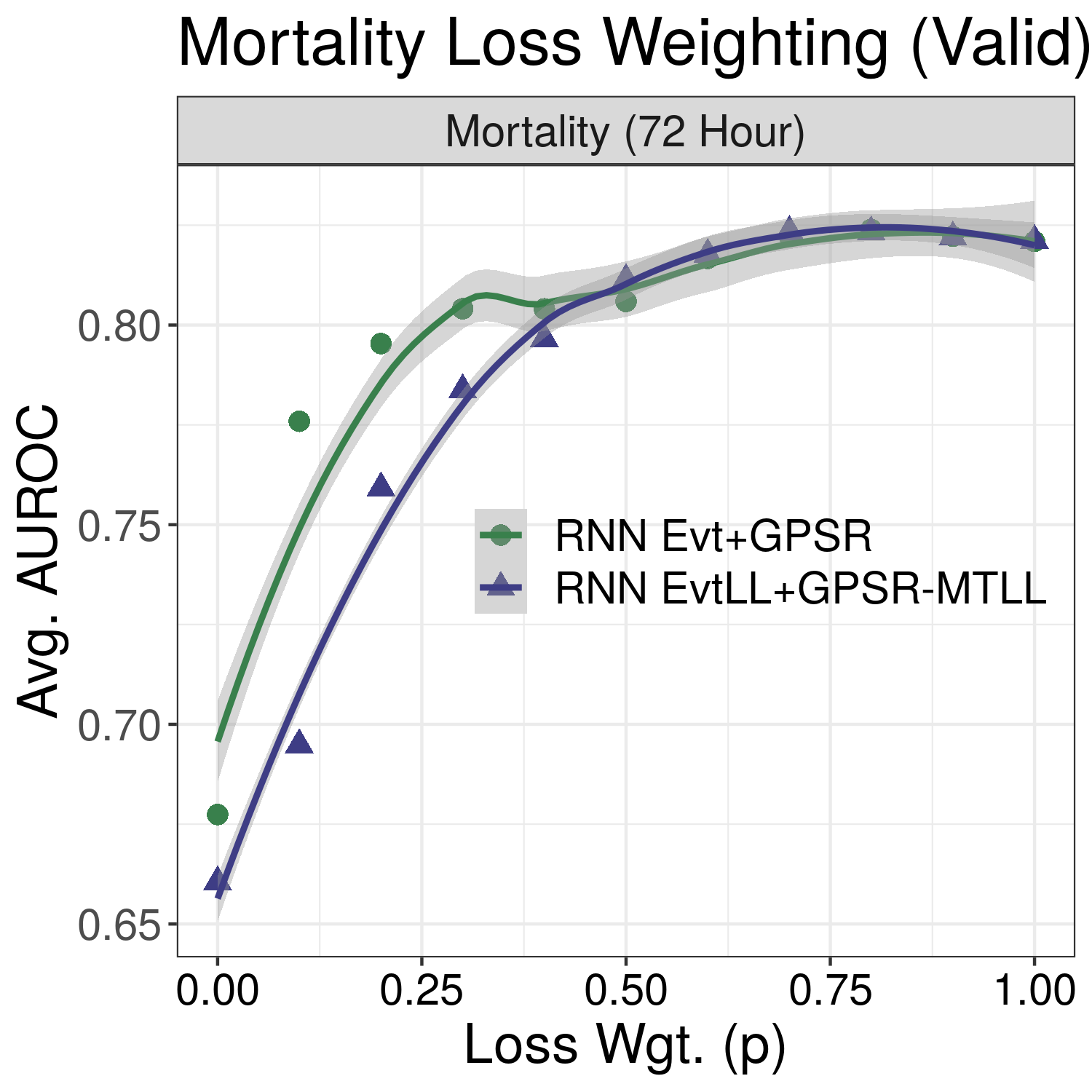}%
}{%
	\vspace{-2mm}
	\caption{Loss weighting search results for mortality validation set AUROC. }
	\label{fig:mortality psearch}
}
\qquad%
\capbtabbox[6cm]{%
	\begin{tabular}{rccc}
	\multicolumn{4}{c}{\textbf{Loss Weighting}}                                                                                                                       \\ \cline{2-4} 
	\multicolumn{1}{c|}{\textbf{Model Name}} & \multicolumn{1}{c|}{\textbf{Mortality}} & \multicolumn{1}{c|}{\textbf{Norepi.}} & \multicolumn{1}{c|}{\textbf{Sepsis}} \\ \hline
	\multicolumn{1}{|r|}{RNN Evt+GPSR}        & \multicolumn{1}{c|}{0.8}                & \multicolumn{1}{c|}{0.8}              & \multicolumn{1}{c|}{0.8}             \\ \hline
	\multicolumn{1}{|r|}{RNN EvtLL+GPSR-MTLL} & \multicolumn{1}{c|}{0.8}                & \multicolumn{1}{c|}{0.9}              & \multicolumn{1}{c|}{0.9}             \\ \hline
	\end{tabular}
}{%
	\centering
	\caption{Best loss weighting results for all targets based on the validation set AUROC.}%
	\label{tbl:p results}
}
\end{floatrow}
\end{figure}

\vspace{-1mm}
Initially, each proposed model structure hyperparameters (e.g. layer sizes) were selected based on AUROC validation performance with a weighted loss of of 0.9, $p=0.9$. After selecting structure hyperparameters, $p$ was iterated over to find the best loss weight according to the validation AUROC. Figure \ref{fig:mortality psearch} demonstrates this search, and Table \ref{tbl:p results} show the selected $p$ for each model and target.

\vspace{-4mm}
\vspace{-12mm}
\section{Results and Discussion}
\vspace{-2mm}
Since we are interested in increasing the performance of low-prior event prediction, the area under the precision-recall curve (AUPRC) is the primary metric used for evaluation. AUROC is also important to ensure that the overall predictive performance is not being heavily sacrificed for better low-prior event prediction, and thus there are additional plots to demonstrate the AUROC performance. Each model is compared under three different conditions (i) average performance, (ii) average performance with a reduced prior likelihood of the positive class, and (iii) average performance over a reduced sample size.
\vspace{-3mm}
\subsection{Predictive Performance}
\begin{figure}
	\centering
	\begin{subfigure}[b]{0.7\textwidth}
	\centering
	\includegraphics[width=\textwidth]{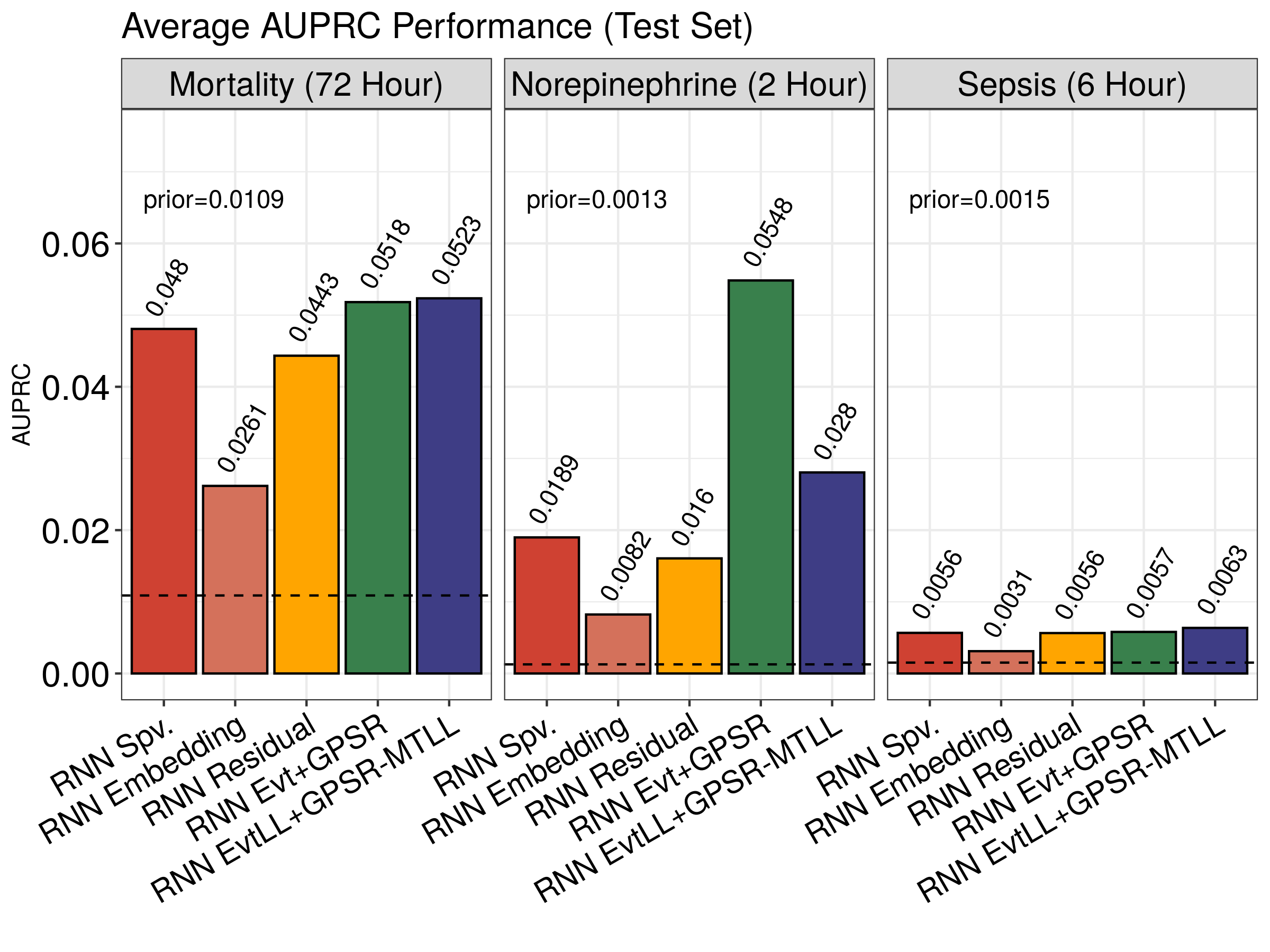}
	\caption{•}
	\vspace{-6mm}
	\label{fig:overall_auprc}
	\end{subfigure}
	\hfill
	\begin{subfigure}[b]{0.8\textwidth}
	\centering
	\includegraphics[width=\textwidth]{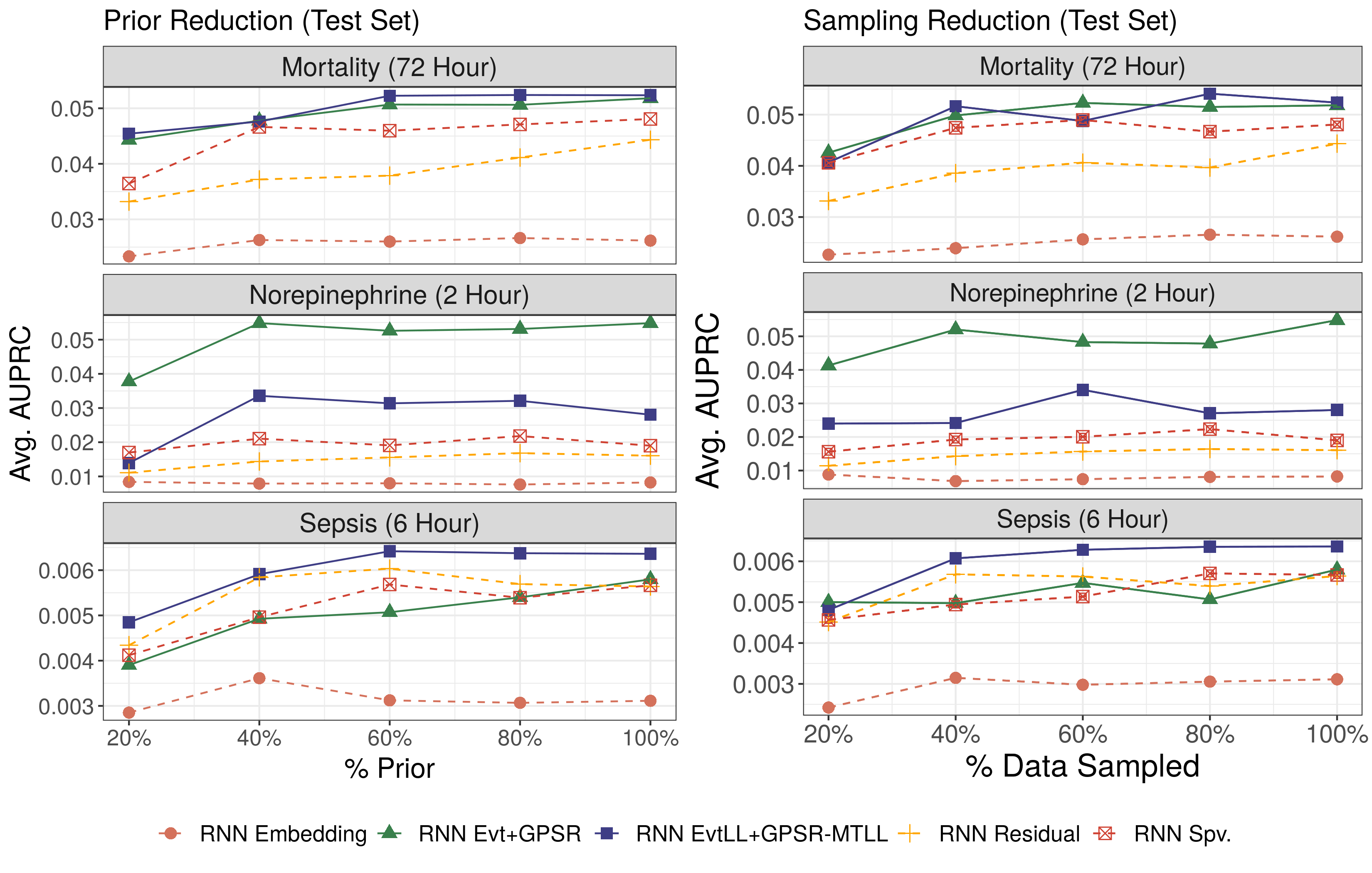}
	\caption{•}
	\label{fig:test_auprc}
	\end{subfigure}
	\vspace{-4mm}
	\caption{\textbf{Figure \ref{fig:overall_auprc}} the average test set AUPRC performance of each proposed and baseline model with a dashed line indicating the test set prior. \textbf{Figure \ref{fig:test_auprc}} (left)the average test set AUPRC when reducing the train/valid prior, and (right) the average test set AUPRC when reducing train/valid samples.}
\end{figure}

\begin{figure}[hbt!]
	\begin{subfigure}[b]{0.45\textwidth}
	\centering
	\includegraphics[width=\textwidth]{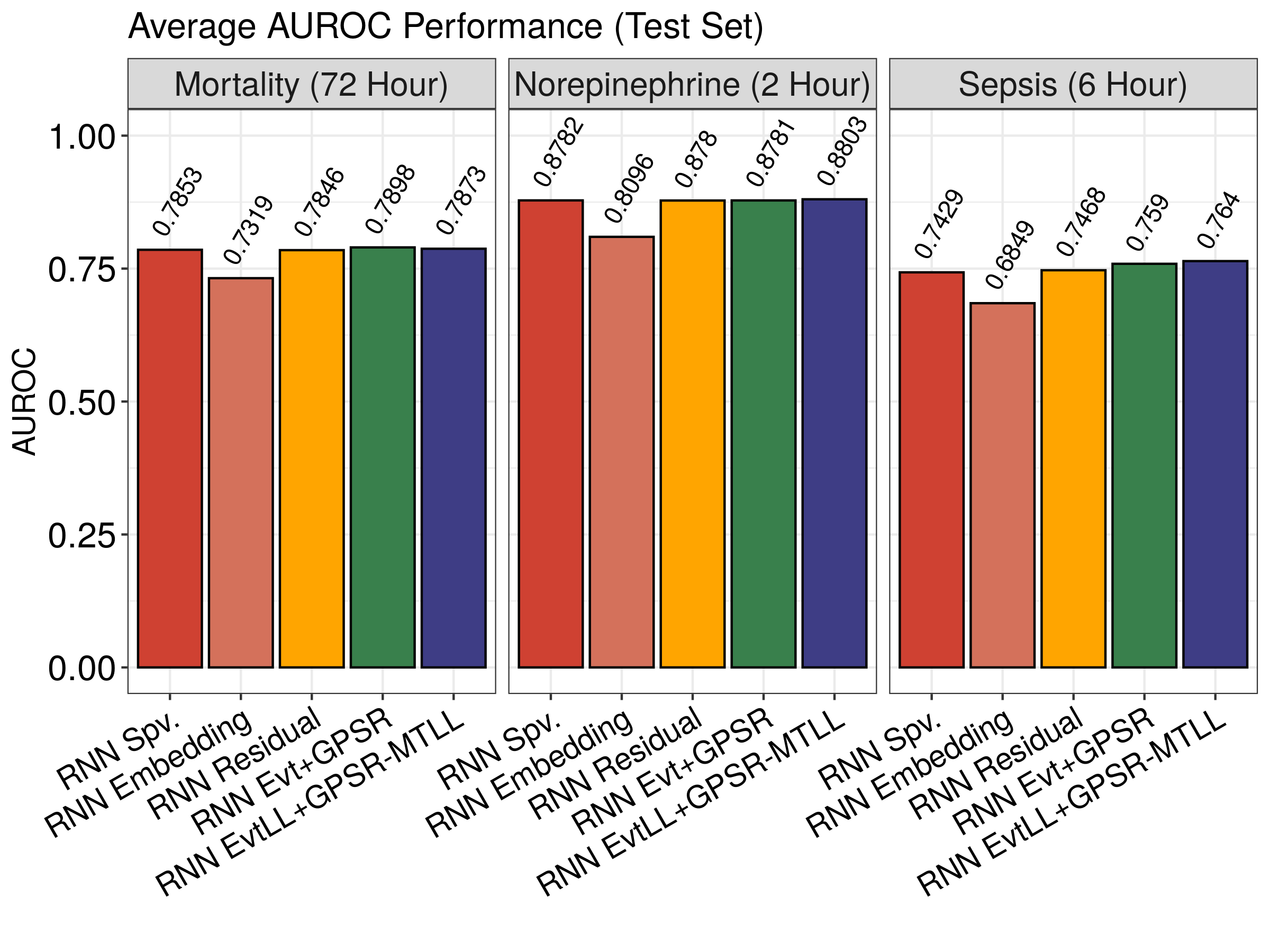}
	\caption{•}
	\label{fig:overall_auroc}
	\end{subfigure}
	\begin{subfigure}[b]{0.52\textwidth}
	\centering
	\includegraphics[width=\textwidth]{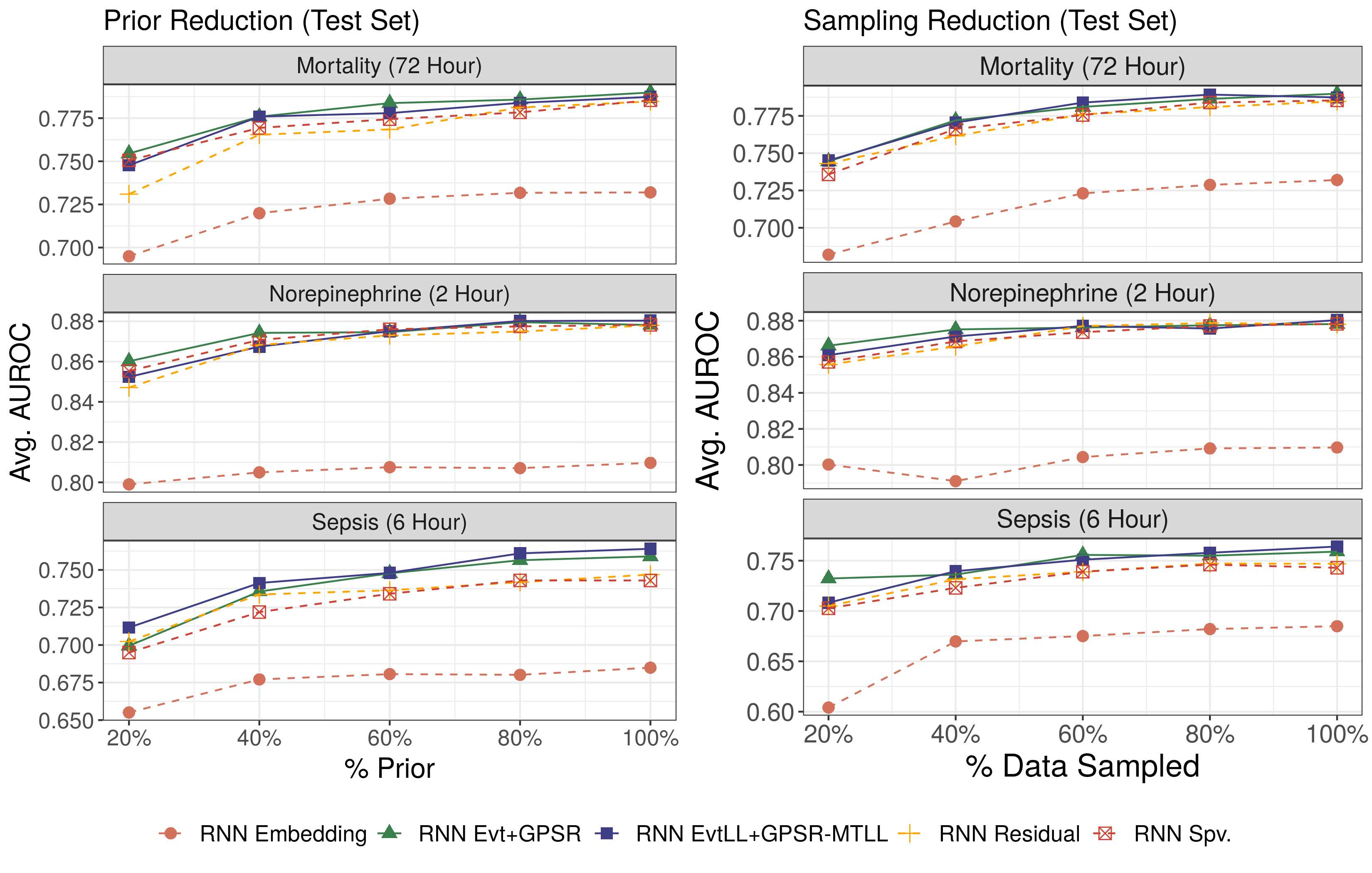}
	\caption{•}
	\label{fig:test_auroc}
	\end{subfigure}
	\vspace{-3mm}
	\caption{Figure \ref{fig:test_auroc} the average test set AUROC performance with reduced prior(left) and samples(right).}
\end{figure}

\vspace{-3mm}
Figure \ref{fig:overall_auprc}, shows an improvement in AUPRC performance over the candidate models when compared to the baselines and the prior likelihood. This performance increase is particularly notable in the prediction of IV Norepinephrine. Figure \ref{fig:overall_auroc} demonstrates that the proposed simultaneous learning models are maintaining a competitive if not stronger AUROC performance compared to the baselines. Based on the results for these three tasks, learning a GPSR simultaneously with a low-prior event provides a competitive to an improved prediction performance. Further, this suggests that low-prior clinical events benefit from the additional signal learning of generic clinical tasks.
\vspace{-4mm}
\vspace{-4mm}
\vspace{4mm}
\subsection{Reduction of Prior Likelihood}
\vspace{-2mm}
By reducing the prior likelihood of each event, the models can be examined as the low-prior prediction target becomes increasingly more challenging to discern. The prior likelihoods of both the training and validation sets were reduced, while the test set remained at the same likelihood. This was performed for 7 iterations of randomly selected positive sequences for each model, and the selections were held constant across models to give a fair comparison. Additionally, the proposed models' loss weight, $p$, was held constant, but hypothetically a more optimal $p$ could have been rendered from this prior reduction.

In Figure \ref{fig:test_auprc}(left), there is a strong AUPRC performance for the majority of prediction events and prior reductions over the two proposed simultaneous learning models. Particularly for norepinephrine and mortality prediction, there is little decrease in AUPRC performance from $100\%$ of the prior to $60\%$. In addition, the AUPRC for the mortality simultaneous models at $20\%$ of the prior is about as strong as the baseline models at $100\%$. This performance holds for the EvtLL+GPSR-MTLL sepsis prediction task where it too maintains a lead on the baseline model performances. Additionally, based on the AUROC Figure \ref{fig:test_auroc}(left), the AUPRC performance does not come with a sacrifice to AUROC. This suggests that the simultaneous learning of low-prior events and general clinical tasks provides support to the prediction of low-prior events even under increasingly sparse conditions.
\vspace{-4mm}
\subsection{Reduction of Sample Size}
\vspace{-2mm}
Sample size reduction provides insight to whether each model is able to be predictive given a more sparse data set. Similar to prior reduction, the iterations of sequence samples are held constant across models to give a fair comparison. The embedding and residual models were given a potential advantage since the GPSR models were not sample reduced.

In Figure \ref{fig:test_auprc}(right), the simultaneous learning models again show strong AUPRC performance. The proposed models on two out of three of the low-prior events have near consistent AUPRC performance up to $40\%$ of the sample size. In addition, the AUROC performance shown in Figure \ref{fig:test_auroc}(right) for the two proposed models maintains a competitive edge over the baselines. Therefore, simultaneous learning of a GPSR to improve low-prior event prediction maintains a competitive edge in a reduced sample data set.
\vspace{-4mm}
\section{Conclusion}
\vspace{-2mm}
Based on the results for these three clinical targets, weighted simultaneous learning of a low-prior event and GPSR improves the prediction of the low-prior task. This prediction improvement is sustained throughout the ablation studies, reduced prior and sample size. This suggests that the predictive signal from forecasting generic clinical tasks provides additional support to the low-prior event, and this predictive benefit can be further capitalized when the low-prior target is simultaneously optimized with the patient representation.
\subsubsection{Acknowledgment.} The work presented was supported by NIH grant R01GM088224. The content of this paper is solely the responsibility of the authors and does not necessarily represent the official views of NIH.
\vspace{-4mm}
\bibliographystyle{splncs04}
\bibliography{splncs04}

\begin{thebibliography}{10}
\providecommand{\url}[1]{\texttt{#1}}
\providecommand{\urlprefix}{URL }
\providecommand{\doi}[1]{https://doi.org/#1}

\bibitem{blei2010supervised}
Blei, D.M., McAuliffe, J.D.: Supervised topic models. arXiv preprint
  arXiv:1003.0783  (2010)

\bibitem{choi2016retain}
Choi, E., Bahadori, M.T., Kulas, J.A., Schuetz, A., Stewart, W.F., Sun, J.:
  Retain: An interpretable predictive model for healthcare using reverse time
  attention mechanism. arXiv preprint arXiv:1608.05745  (2016)

\bibitem{gers2000learning}
Gers, F.A., Schmidhuber, J., Cummins, F.: Learning to forget: Continual
  prediction with lstm. Neural computation  \textbf{12}(10),  2451--2471 (2000)

\bibitem{gupta2020transfer}
Gupta, P., Malhotra, P., Narwariya, J., Vig, L., Shroff, G.: Transfer learning
  for clinical time series analysis using deep neural networks. Journal of
  Healthcare Informatics Research  \textbf{4}(2),  112--137 (2020)

\bibitem{gupta2018using}
Gupta, P., Malhotra, P., Vig, L., Shroff, G.: Using features from pre-trained
  timenet for clinical predictions. In: KHD@ IJCAI (2018)

\bibitem{hauskrecht2016outlier}
Hauskrecht, M., Batal, I., Hong, C., Nguyen, Q., Cooper, G.F., Visweswaran, S.,
  Clermont, G.: Outlier-based detection of unusual patient-management actions:
  an icu study. Journal of biomedical informatics  \textbf{64},  211--221
  (2016)

\bibitem{hauskrecht2013outlier}
Hauskrecht, M., Batal, I., Valko, M., Visweswaran, S., Cooper, G.F., Clermont,
  G.: Outlier detection for patient monitoring and alerting. Journal of
  biomedical informatics  \textbf{46}(1),  47--55 (2013)

\bibitem{johnson2016mimic}
Johnson, A.E., Pollard, T.J., Shen, L., Li-Wei, H.L., Feng, M., Ghassemi, M.,
  Moody, B., Szolovits, P., Celi, L.A., Mark, R.G.: Mimic-iii, a freely
  accessible critical care database. Scientific data  \textbf{3}(1), ~1--9
  (2016)

\bibitem{kratz2004laboratory}
Kratz, A., Ferraro, M., Sluss, P.M., Lewandrowski, K.B.: Laboratory reference
  values. New England Journal of Medicine  \textbf{351},  1548--1564 (2004)

\bibitem{laposata2019laposata}
Laposata, M.: Laposata's Laboratory Medicine Diagnosis of Disease in Clinical
  Laboratory Third Edition. McGraw-Hill Education (2019)

\bibitem{lee2021modeling}
Lee, J.M., Hauskrecht, M.: Modeling multivariate clinical event time-series
  with recurrent temporal mechanisms. Artificial Intelligence in Medicine
  \textbf{112},  102021 (2021)

\bibitem{lei2018effective}
Lei, L., Zhou, Y., Zhai, J., Zhang, L., Fang, Z., He, P., Gao, J.: An effective
  patient representation learning for time-series prediction tasks based on
  ehrs. In: 2018 IEEE International Conference on Bioinformatics and
  Biomedicine (BIBM). pp. 885--892. IEEE (2018)

\bibitem{loshchilov2017decoupled}
Loshchilov, I., Hutter, F.: Decoupled weight decay regularization. arXiv
  preprint arXiv:1711.05101  (2017)

\bibitem{lyu2018improving}
Lyu, X., Hueser, M., Hyland, S.L., Zerveas, G., Raetsch, G.: Improving clinical
  predictions through unsupervised time series representation learning. arXiv
  preprint arXiv:1812.00490  (2018)

\bibitem{malakouti2019predicting}
Malakouti, S., Hauskrecht, M.: Predicting patient’s diagnoses and diagnostic
  categories from clinical-events in ehr data. In: Conference on Artificial
  Intelligence in Medicine in Europe. pp. 125--130. Springer (2019)

\bibitem{mcdonald2003loinc}
McDonald, C.J., Huff, S.M., Suico, J.G., Hill, G., Leavelle, D., Aller, R.,
  Forrey, A., Mercer, K., DeMoor, G., Hook, J., et~al.: Loinc, a universal
  standard for identifying laboratory observations: a 5-year update. Clinical
  chemistry  \textbf{49}(4),  624--633 (2003)

\bibitem{miotto2016deep}
Miotto, R., Li, L., Kidd, B.A., Dudley, J.T.: Deep patient: an unsupervised
  representation to predict the future of patients from the electronic health
  records. Scientific reports  \textbf{6}(1),  1--10 (2016)

\bibitem{rajkomar2018scalable}
Rajkomar, A., Oren, E., Chen, K., Dai, A.M., Hajaj, N., Hardt, M., Liu, P.J.,
  Liu, X., Marcus, J., Sun, M., et~al.: Scalable and accurate deep learning
  with electronic health records. NPJ Digital Medicine  \textbf{1}(1),  1--10
  (2018)

\bibitem{ren2019prediction}
Ren, J., Kunes, R., Doshi-Velez, F.: Prediction focused topic models for
  electronic health records. arXiv preprint arXiv:1911.08551  (2019)

\bibitem{reyna2019early}
Reyna, M.A., Josef, C., Seyedi, S., Jeter, R., Shashikumar, S.P., Westover,
  M.B., Sharma, A., Nemati, S., Clifford, G.D.: Early prediction of sepsis from
  clinical data: the physionet/computing in cardiology challenge 2019. In: 2019
  Computing in Cardiology (CinC). pp. Page--1. IEEE (2019)

\bibitem{smith2019norepinephrine}
Smith, M.D., Maani, C.V.: Norepinephrine. StatPearls [Internet]  (2019)

\bibitem{tomavsev2019clinically}
Toma{\v{s}}ev, N., Glorot, X., Rae, J.W., Zielinski, M., Askham, H., Saraiva,
  A., Mottram, A., Meyer, C., Ravuri, S., Protsyuk, I., et~al.: A clinically
  applicable approach to continuous prediction of future acute kidney injury.
  Nature  \textbf{572}(7767),  116--119 (2019)

\end{thebibliography}
\vspace{-6mm}
\end{document}